\documentclass[11pt]{article}

\usepackage{acl}

\usepackage{times}
\usepackage{latexsym}

\usepackage[T1]{fontenc}

\usepackage[utf8]{inputenc}

\usepackage{microtype}

\usepackage{inconsolata}

\usepackage{graphicx}
\usepackage{latexsym}
\usepackage[T1]{fontenc}
\usepackage[utf8]{inputenc}
\usepackage{microtype}
\usepackage{booktabs}
\usepackage{threeparttable}
\usepackage{stfloats}
\usepackage{array}
\usepackage{tabularx}
\usepackage{fvextra}

\usepackage[main=english,ukrainian]{babel}

\newcolumntype{L}[1]{>{\raggedright\arraybackslash}p{#1}}
\newcolumntype{Y}{>{\raggedright\arraybackslash}X}

\newcommand{\promptlabel}[1]{\noindent\textit{#1}\par}
\DefineVerbatimEnvironment{Prompt}{Verbatim}
  {fontsize=\footnotesize,breaklines=true,breaksymbolleft={},fontfamily=rm}

\title{Qwen Goes Brrr: Off-the-Shelf RAG for Ukrainian Multi-Domain Document Understanding}

\author{Anton Bazdyrev \ Ivan Bashtovyi \ Ivan Havlytskyi \\ {\bf Oleksandr Kharytonov} \ {\bf Artur Khodakovskyi} \ \\ 
National Technical University of Ukraine "Igor Sikorsky Kyiv Polytechnic Institute"}

\begin{document}
\maketitle

\begin{abstract}
We participated in the Fifth UNLP shared task on multi-domain document understanding, where systems must answer Ukrainian multiple-choice questions from PDF collections and localize the supporting document and page. We propose a retrieval-augmented pipeline built around three ideas: contextual chunking of PDFs, question-aware dense retrieval and reranking conditioned on both the question and answer options, and constrained answer generation from a small set of reranked passages. Our final system uses Qwen3-Embedding-8B for retrieval, a fine-tuned Qwen3-Reranker-8B for passage ranking, and Qwen3-32B for answer selection. On a held-out split, reranking improves Recall@1 from 0.6957 to 0.7935, while using the top-2 reranked passages raises answer accuracy from 0.9348 to 0.9674. Our best leaderboard run reached 0.9452 on the public leaderboard and 0.9598 on the private leaderboard. Our results suggest that, under strict code-competition constraints, preserving document structure and making relevance estimation aware of the answer space are more effective than adding complex downstream heuristics.
\end{abstract}

\section{Introduction}
\subsection{Motivation \& Context}
Real-world document understanding goes beyond extracting an answer span from a passage. A system must navigate long, heterogeneous, domain-specific documents, locate the right evidence among distractors, and tie that evidence to a concrete decision. Generative models alone tend to hallucinate when grounding is required, while retrieval-only pipelines lose critical context once chunks are stripped from their surrounding document structure.

The UNLP 2026 shared task \cite{sydorskyi-etal-2026-unlp} makes these difficulties concrete. Submissions must produce three coupled outputs --- the correct multiple-choice answer, the source document, and the page grounding the answer --- under scarce task-specific training data, diverse PDF formatting, and strict runtime budgets imposed by the code-competition setting.

\subsection{Shared Task Overview}
The shared task is organized as a Kaggle code competition. Given a set of PDF documents and a six-option multiple-choice question, the system must predict: (1) the correct answer from options A--F, (2) the supporting document, and (3) the supporting page. The visible training data contain two domains, while the hidden submission set additionally includes an unseen \texttt{domain\_3}. The visible test set is only a dummy subset; at submission time it is replaced with a substantially larger hidden set of approximately 240 PDFs. All solutions must run within nine hours on Kaggle hardware without internet access.

This setup strongly favors efficient multi-stage pipelines. The system must maintain high retrieval recall under domain shift, but it must also keep the final answerer small enough to fit the competition runtime budget.

\subsection{Contributions}
We make three main contributions:
\begin{enumerate}
\item We present a state-of-the-art retrieval-augmented generation pipeline for Ukrainian multi-domain document understanding that combines structure-aware chunking, question-aware retrieval, and answer-aware reranking within the practical constraints of a code-only shared-task setting.
\item We show that highly competitive performance can be achieved with largely off-the-shelf pretrained models, and that careful pipeline design---particularly preserving document structure and conditioning relevance estimation on the full multiple-choice instance---matters more than adding complex downstream heuristics.
\item We construct new Ukrainian resources for retrieval-based multiple-choice question answering, including training data for both dense retrieval and reranking.
\end{enumerate}

\section{Related Work}

\subsection{UA-SQuAD}
UA-SQuAD is a key resource for Ukrainian question answering, providing supervised data for reading comprehension in a language with comparatively limited task-specific QA benchmarks \cite{ua_datasets_2021}. More broadly, extractive QA datasets such as SQuAD and its multilingual extensions have been central for training and evaluating models that align questions with answer-bearing passages, and they often serve as useful transfer supervision even beyond purely extractive settings \cite{rajpurkar-etal-2016-squad,rajpurkar-etal-2018-know}. In this sense, UA-SQuAD is relevant not only as a benchmark, but also as a practical source of supervision for Ukrainian retrieval and evidence-selection pipelines.

\subsection{Document Processing and Contextual Chunking}
Prior work on long-document understanding has shown that document processing and segmentation strongly influence retrieval and downstream QA quality, especially for PDFs and other structured sources where headings, layout, and local reading order carry important information \cite{wang-etal-2025-document, lin2024chatdoc}. While fixed-size chunking is a common baseline, more structure-aware and context-preserving chunk construction can yield better retrieval units by retaining document titles, section paths, or neighboring context, which makes passages more self-contained for both ranking and answering \cite{chen-etal-2024-dense, singh2025contextual}. This line of work motivates contextual chunking as a natural design choice in retrieval-augmented document QA \cite{hichunk2025}.

\subsection{Retrieval in Ukrainian Language}
Retrieval for QA is commonly built from dense retrievers, rerankers, and answer modules, with multilingual and cross-lingual transfer playing a major role when native-language retrieval supervision is scarce \cite{wang-etal-2024-retrieve, limkonchotiwat-etal-2024-mccrolin}. This is particularly relevant for Ukrainian, where retrieval resources remain more limited than in English, making adaptation from multilingual models and QA-derived supervision especially important \cite{haltiuk-smywinski-pohl-2025-path}. For evidence-based QA, retrieval is not only a search problem but also a passage selection problem, which is why strong reranking and evidence filtering are often necessary in addition to first-stage retrieval \cite{limkonchotiwat-etal-2024-mccrolin}.

\section{Dataset}

\subsection{Data Source \& Task Format}
The official competition data was provided in two releases: an initial training set of 40 questions and a subsequently released development set of 461 questions. Both sets are grounded in the same 41 PDF documents spanning two visible domains. Each question has exactly six answer options and is annotated with the correct option, the supporting document, and the supporting page. The hidden submission set includes a third unseen domain, which makes generalization across domains central to system design.

The underlying documents are not short passages but full PDFs with varied formatting, section hierarchies, tables, and long local dependencies. This makes page-level retrieval non-trivial even when document-level identification is relatively easy.

\subsection{Auxiliary Training Data}
To fine-tune our models, we used a combined dataset that merges competition supervision with auxiliary Ukrainian extractive QA data. We specifically leveraged UA-SQuAD \cite{ua_datasets_2021}, a Ukrainian adaptation of the widely used SQuAD 2.0 benchmark \cite{rajpurkar-etal-2018-know}. This dataset provides paragraph-level contexts from Wikipedia paired with localized questions. Incorporating this data allows the model to learn fine-grained context comprehension and exact answer localization before adapting to the longer, noisier PDFs in the competition domain. The final reranker training set contains:
\begin{itemize}
\item 461 competition examples;
\item 16,658 answerable UA-SQuAD question-context pairs;
\item hard negatives mined from impossible UA-SQuAD questions, dense retrieval errors, confusing wrong-answer passages, and same-document wrong-page passages.
\end{itemize}

\subsection{Large-Scale Retrieval Pretraining Corpus}
To address the scarcity of high-quality retrieval datasets for the Ukrainian language, we constructed a novel 80,000-example pretraining corpus.\footnote{\url{https://huggingface.co/datasets/G37A/ua-retrieval-80k-silver}} Although our final, strongest pipeline did not strictly require these additional pretraining signals, this dataset represents a significant standalone resource. To our knowledge, it is the first large-scale dataset explicitly designed for training bi-encoder embedding models and cross-encoder rerankers in Ukrainian.

We aggregated queries and passages from four established English retrieval datasets to ensure broad domain coverage. The corpus consists of 30,000 rows from Natural Questions \cite{aarsen_natural_questions_hard_negatives}, 20,000 from HotpotQA \cite{yang2018hotpotqa}, 20,000 from MS MARCO \cite{DBLP:journals/corr/NguyenRSGTMD16}, and 10,000 from GooAQ \cite{gooaq2021}. This composition captures a wide range of text styles, from formal encyclopedia articles and multi-hop reasoning contexts to everyday web searches and layperson explanations. Overall, the queries range from 5 to 630 characters (with a median of 45), while the positive passages vary significantly from 10 to 9,437 characters (with a median of 372).

We made several deliberate design choices to ensure the dataset is highly effective for contrastive learning:
\begin{itemize}
\item \textbf{Original Pre-mined Hard Negatives:} Rather than mining new negatives post-translation, we explicitly preserved the original hard negatives provided within the source datasets. Every row includes up to three hard negatives originally pre-mined via dense retrievers or BM25 by the respective dataset creators, forcing the models to learn subtle semantic differences while avoiding translation-induced mining artifacts.
\item \textbf{Passage Length Diversity:} The extreme variance in document length ensures the resulting models become robust to different chunking strategies, whether they are processing tight 512-token windows or full-page contexts.
\item \textbf{Cross-Domain Generalization:} Combining diverse domains helps build a generalized retrieval mechanism, reducing the risk of overfitting before task-specific adaptation.
\end{itemize}

Because the source material is in English, the entire corpus is translated into Ukrainian. To maintain high semantic accuracy during translation, we utilized high-capacity, language-optimized model: our self-hosted TranslateGemma-27B \cite{translategemma2026}. The dataset is distributed as a Parquet file natively compatible with standard contrastive learning frameworks.

Table~\ref{tab:data} summarizes the resulting training sources, the number of positive and negative examples, and the corresponding negative-mining strategies.

\begin{table*}[t]
\small
\centering
\setlength{\tabcolsep}{6pt}
\begin{tabularx}{\textwidth}{@{}l r r X X@{}}
\toprule
\textbf{Dataset} & \textbf{Positives} & \textbf{Negatives} & \textbf{Negative Mining} & \textbf{Original Source} \\
\midrule
Competition train & 40 & 200 & Ground truth (wrong options) & 41 PDFs (2 visible domains) \\
Competition dev & 461 & 2,305 & Ground truth (wrong options) & Same 41 PDFs (Released later) \\
UA-SQuAD & 16,658 & 5,766 & Impossible Qs \& dense errors & Ukrainian Extractive QA \\
Retrieval Pretraining & 80,000 & $\sim$240,000 & Pre-mined (BM25 \& Dense) & NQ, HotpotQA, MS MARCO, GooAQ \\
\bottomrule
\end{tabularx}
\caption{Dataset composition detailing the volume of positive and negative examples. The mining strategy column highlights how contrastive pairs were sourced, while the original source column indicates the underlying documents or datasets.}
\label{tab:data}
\end{table*}

\section{Evaluation Metric and Competition Constraints}
The competition metric combines answer correctness and reference quality, where reference quality includes both document identification and page localization. For a test set of $N$ questions, the final score is defined as
\[
\mathrm{Score}
=
\frac{1}{2} \cdot \frac{1}{N}\sum_{i=1}^{N} a_i
+
\frac{1}{4} \cdot \frac{1}{N}\sum_{i=1}^{N} d_i
+
\frac{1}{4} \cdot \frac{1}{N}\sum_{i=1}^{N} p_i,
\]
where
\[
a_i = \mathbf{1}\!\left(\hat{y}_i = y_i\right)
\]
indicates whether the predicted answer $\hat{y}_i$ matches the gold answer $y_i$, and
\[
d_i = \mathbf{1}\!\left(\widehat{\mathrm{Doc}}_i = \mathrm{Doc}_i\right)
\]
indicates whether the predicted supporting document matches the gold document.

The page-level component is defined as
\[
p_i
=
\left(
1 - \frac{\left|\widehat{\mathrm{Page}}_i - \mathrm{Page}_i\right|}{n_i}
\right)
\mathbf{1}(d_i = 1),
\]
where $\widehat{\mathrm{Page}}_i$ is the predicted page, $\mathrm{Page}_i$ is the gold page, and $n_i$ is the number of pages in the gold document. Thus, page proximity is credited only when the predicted document is correct; otherwise, the page contribution is zero.

This metric has an important modeling consequence. The system does not need only to retrieve relevant evidence; it must retrieve evidence that is both answer-discriminative and localized tightly enough to preserve page-level precision. Because the hidden test set is much larger than the visible one and includes an unseen domain, explicit domain probing is forbidden and brittle domain-specific routing is risky.

\section{Method}

\subsection{Technical Details}
Retrieval and reranking were run with vLLM-based inference, while the final answerer used constrained single-token generation over the answer alphabet. All major design choices were made under the practical requirement that the complete pipeline fit within the Kaggle code-competition budget on 2$\times$T4 hardware.\footnote{\url{https://github.com/nuinashco/unlp2026_shared_task}}

Our development process followed a simple progression. We first stabilized chunking, then improved first-stage recall, then focused on reranking, and only after that scaled the final answerer. This ordering turned out to be important: a stronger generator alone could not compensate for structurally weak evidence.

\subsection{Contextual Chunking}
Our first design decision was to avoid treating a PDF page as plain text. Instead, we chunk documents in a structure-aware way.

\paragraph{PDF processing.}
We initially used Docling \cite{Docling}, a state-of-the-art document understanding library offering layout analysis and table structure recognition. However, Docling proved both slow and unstable in the competition environment, failing on a non-trivial fraction of PDFs. We replaced it with a custom chunker built on \texttt{pymupdf4llm}, which converts PDF pages to Markdown via fast native rendering. This replacement yielded a substantial speedup with no drop in leaderboard score.

\paragraph{Chunk structure.}
Each chunk contains three nested levels of context:
\begin{enumerate}
\item a short document prefix (global context extracted from the document start);
\item the current heading path (section context);
\item the local chunk body.
\end{enumerate}

In the retained configuration, chunks are built with a maximum length of 512 tokens, 64-token overlap, and a 128-token document prefix. On the 41 competition training documents, this produced 3,362 contextual chunks.

\subsection{Dense Retrieval}
 
We use dense bi-encoder retrieval over contextualized chunks. Each query is built from the full multiple-choice instance: the question, answer options, and a short instruction emphasizing document and page grounding. The retriever returns the top-20 chunks for reranking. The exact prompt is provided in \autoref{fst:appendix}.
 
We compared off-the-shelf and fine-tuned embedding models of different sizes; the results are shown in Table~\ref{tab:embedding_experiments}. The pretrained \texttt{Qwen3 8B} embedder \cite{zhang2025qwen3embeddingadvancingtext} gave the best overall retrieval backbone and remained strongest without task-specific adaptation. Fine-tuning it on UA-SQuAD brought no gain, suggesting that larger embedding models already generalize well in this setting and overfit quickly on narrow supervision.
 
Fine-tuning was more useful for smaller models. For \texttt{Qwen3 0.6B}, training on UA-SQuAD and our \texttt{80k} corpus substantially reduced the gap to the pretrained \texttt{8B} model, while adding UNLP development data degraded performance. We also tested the off-the-shelf diffusion-style embedding model \texttt{pplx-embed-v1-4b} \cite{eslami2026diffusionpretraineddensecontextualembeddings}, but it performed markedly worse than the Qwen3-based alternatives.
 
Overall, the results indicate a practical trade-off: if compute allows, a larger pretrained embedder is the strongest choice; under tighter resource limits, fine-tuning a smaller model can recover much of the gap.
 
\begin{table}[t]
\centering
\small
\begin{threeparttable}
\setlength{\tabcolsep}{5pt}
\begin{tabular}{lcc}
\toprule
\textbf{Model} & \textbf{Public} & \textbf{Private} \\
\midrule
Qwen3-Embedding-8B & 0.9426 & 0.9592 \\
Qwen3-Embedding-8B + squad & 0.9414 & 0.9591 \\
\midrule
Qwen3-Embedding-0.6B + 80k + squad & 0.9390 & 0.9581 \\
Qwen3-Embedding-0.6B + squad & 0.9345 & 0.9534 \\
Qwen3-Embedding-0.6B & 0.9370 & 0.9507 \\
Qwen3-Embedding-0.6B + full\tnote{*} & 0.9289 & 0.9427 \\
Qwen3-Embedding-0.6B + 80k & 0.9281 & 0.9400 \\
\midrule
pplx-embed-v1-4b & 0.8737 & 0.8759 \\
\bottomrule
\end{tabular}
\begin{tablenotes}
  \small
  \item[*] 80k + squad + unlp
\end{tablenotes}
\end{threeparttable}
\caption{Embedding comparison with fixed off-the-shelf \texttt{Qwen3-Reranker-8B} and \texttt{Qwen3-32B-AWQ} generator \cite{yang2025qwen3technicalreport}. \texttt{80k} denotes our machine-translated auxiliary retrieval corpus.}
\label{tab:embedding_experiments}
\end{table}
 
\subsection{Option-Aware Reranking}
 
The second stage of our pipeline employs Qwen3-Reranker over the top-20 chunks returned by the dense retriever. Crucially, the reranker receives the question concatenated with all six answer options, rather than the question in isolation. This structural change proved decisive: for multiple-choice QA, passage usefulness often depends on which specific alternatives must be distinguished, not just on broad topical similarity. The exact system and user prompts are provided in \autoref{sec:appendix}.
 
To train the model for this discriminative behavior, we relied on the hard negatives detailed in our dataset formulation, specifically emphasizing retrieved chunks that support plausible but incorrect options. After scoring all candidate chunks, we retain only the top-2 passages to pass forward to the final answer generator.
 
We evaluated reranker models across several sizes and fine-tuning configurations; the results are shown in Table~\ref{tab:reranker_experiments}. The \texttt{Qwen3-Reranker-8B} model fine-tuned exclusively on the auxiliary UA-SQuAD dataset (\texttt{+ squad}) provided the strongest overall performance on the Private leaderboard.
 
As with the embedding stage, the large 8B reranker proved susceptible to overfitting when exposed to competition-specific data. Introducing the UNLP development set into the fine-tuning mix (\texttt{+ unlp} and \texttt{+ squad + unlp}) noticeably degraded performance, suggesting the model overfit to the characteristics of the two visible domains. Conversely, combining UA-SQuAD with the diverse 80k corpus (\texttt{+ squad + 80k}) avoided this degradation, achieving the highest Public score while remaining highly competitive on the unseen domain.
 
For smaller models (\texttt{4B} and \texttt{0.6B}), the off-the-shelf versions performed reasonably well but consistently degraded when fine-tuned in any configuration, suggesting that smaller rerankers lack the capacity to learn the complex question-and-all-options formatting without catastrophic forgetting of their general retrieval capabilities.
 
\begin{table}[t]
\centering
\small
\setlength{\tabcolsep}{5pt}
\begin{tabular}{lcc}
\toprule
\textbf{Model} & \textbf{Public} & \textbf{Private} \\
\midrule
Qwen3-Reranker-8B & 0.9426 & 0.9592 \\
Qwen3-Reranker-8B + unlp & 0.9286 & 0.9477 \\
Qwen3-Reranker-8B + squad & 0.9452 & \textbf{0.9598} \\
Qwen3-Reranker-8B + squad + unlp & 0.9330 & 0.9514 \\
Qwen3-Reranker-8B + 80k + squad & \textbf{0.9456} & 0.9590 \\
\midrule
Qwen3-Reranker-4B & 0.9429 & 0.9562 \\
Qwen3-Reranker-4B + squad & 0.9253 & 0.9519 \\
\midrule
Qwen3-Reranker-0.6B & 0.9172 & 0.9395 \\
Qwen3-Reranker-0.6B + squad & 0.9007 & 0.9257 \\
Qwen3-Reranker-0.6B + squad + unlp & 0.9082 & 0.9360 \\
\midrule
\textit{No Reranker (Baseline)} & 0.9099 & 0.9243 \\
\bottomrule
\end{tabular}
\caption{Reranker comparison with fixed off-the-shelf \texttt{Qwen3-Embedding-8B} retriever and \texttt{Qwen3-32B-AWQ} generator. The Private leaderboard evaluates on an unseen third domain.}
\label{tab:reranker_experiments}
\end{table}

\subsection{Answer Selection and Localization}

The final answer is produced by Qwen3-32B AWQ quantized \cite{lin2024awqactivationawareweightquantization} from the question, six answer options, and a small set of top reranked passages. We expose passage rank in the prompt so that lower ranks indicate stronger relevance, and constrain decoding to a single token from A--F, which makes inference stable and efficient. The predicted document and page are taken from the highest-ranked selected chunk. The exact system and user prompts are provided in \autoref{trd:appendix}.

We also tested a confidence-based variant that adaptively passed one, two, or three passages to the generator based on the top reranker score. Although this was conceptually appealing, it performed slightly worse than the simpler fixed evidence-packing strategy, which we therefore retained.

\subsection{Results}

Table~\ref{tab:ablations} summarizes key experiments on the competition leaderboard.

The largest single gain comes from document-context prepending: removing it drops the public score from 0.945 to 0.861, confirming that structural signals are critical for passage disambiguation. Replacing the prepended document context with an 80-token summary of the document's first 1024 tokens also hurts: the compressed summary loses the structural signals that make raw prepending effective. Conditioning retrieval and reranking on the full multiple-choice instance, together with reranker fine-tuning, pushes the final system to \textbf{0.9598} on the private leaderboard.

To support reproducibility, we release the trained checkpoints and auxiliary resources in a public Hugging Face collection.\footnote{\url{https://huggingface.co/collections/G37A/qwen-goes-brrr}}

\begin{table}[h]
\centering
\small
\begin{threeparttable}
\begin{tabular}{lcc}
\toprule
Variant & Public & Private \\
\midrule
Generator only (no RAG)\tnote{*} & 0.3352 & 0.3392 \\
Early system (BM25, Qwen2.5-32B) & 0.9035 & 0.9114 \\
\midrule
\multicolumn{3}{l}{\textit{Ablations from the final pipeline}} \\
\midrule
w/o document-context prepending & 0.8610 & 0.8891 \\
w/ summary injection (1024$\to$80) & 0.9177 & 0.9408 \\
14B generator, top-10$\to$1 & 0.9346 & 0.9483 \\
Rerank Q+A only (retrieval: Q alone) & 0.9397 & 0.9570 \\
Baseline & 0.9426 & 0.9592 \\
\midrule
\textbf{Final Pipeline (fine-tuned reranker)} & \textbf{0.9452} & \textbf{0.9598} \\
\bottomrule
\end{tabular}
\begin{tablenotes}
  \small
  \item[*] Retrieval score set to 0 (no retrieval component); scores are effectively out of 0.5.
\end{tablenotes}
\end{threeparttable}
\caption{Leaderboard scores for key variants. Ablation rows use Qwen3-Embedding-8B, Qwen3-Reranker-8B, top-20$\to$2 selection, full Q+A at both retrieval and reranking, and Qwen3-32B-AWQ, unless otherwise noted.}
\label{tab:ablations}
\end{table}

\section{Alternative Approaches}

\subsection{Gemma3-based models and bidirectionality}
We considered a Gemma3-based \cite{gemmateam2025gemma3technicalreport, MamayLMv1, Paniv_Lapa_LLM_v0_1_2_2025} alternative motivated in part by our earlier work \cite{bazdyrev-etal-2025-transforming} on adapting decoder-only models toward bidirectional encoding behavior for downstream understanding tasks. That line of work is relevant here because bidirectionality is appealing for evidence retrieval and page-level localization, where useful signals may depend on relationships spanning both left and right context. More broadly, tasks such as retrieval and named entity recognition often benefit from bidirectional representations when fine-grained contextual interactions are important.

We did not retain this direction in the final system for two practical reasons. First, introducing bidirectional computation substantially increases inference and adaptation cost, which is difficult to accommodate under the strict runtime limits of the shared task. Second, Gemma3 is less convenient in the target hardware setting: Tesla T4 GPUs do not support BF16, while FP16 deployment is known to be unstable for Gemma3 \cite{han2025gemma3}, and maintaining higher-precision inference is too expensive under the competition budget. We therefore prioritized a Qwen-based pipeline with a better efficiency--quality trade-off for the code-only setting. At the same time, we view bidirectional adaptation of decoder models as a promising direction for tasks where bidirectional dependencies are central and compute constraints are less restrictive.

\subsection{Diffusion and Contextual Embeddings}
We also explored diffusion-based embedding models inspired by recent work on diffusion-pretrained dense and contextual retrieval embeddings. These models are attractive because they support bidirectional attention, and the contextual variant additionally incorporates document-level context into passage representations.

In our experiments, this direction underperformed the retained Qwen3-based retrieval setup. The off-the-shelf diffusion embedder \texttt{pplx-embed-v1-4b} scored well below the best Qwen3 models in the end-to-end pipeline, and the contextual diffusion variant \texttt{pplx-embed-context-v1} also performed worse on validation. Even so, we consider this line promising: bidirectional modeling and stronger context handling remain appealing for retrieval, and Ukrainian pretraining with in-domain fine-tuning may yield better results in future work.

\subsection{Summary in Context}
Rather than prepending the raw first 128 tokens of a document to each chunk — which can be noisy and may truncate before reaching meaningful content — we explored replacing that prefix with a generated summary of the document's leading section (first 1024, 2048, or 4096 tokens). The hypothesis was that a compact, model-generated summary would expose document-level semantics more reliably than a fixed token cut. We used Qwen3-8B-AWQ to generate these summaries. However, this approach degraded retrieval quality: using the top-1024 token summary, public leaderboard score dropped from 0.9346 to 0.9177.

\subsection{Domain Classifier}
We added a domain-classifier-first routing stage in which each question was assigned to a predicted domain and retrieval was restricted to the corresponding document partition. Domain labels were obtained by prompting the LLM on each folder's readme, while question domains were classified via a zero-shot prompt. To avoid additional model loads, domain masking was applied directly to the embedding similarity matrix: scores for out-of-domain chunks were set to $-\infty$ before top-k selection, preserving retrieval latency.

On the public leaderboard this configuration achieved our highest score, as the clean separation between the two visible domains allowed the classifier to operate with near-perfect accuracy. However, the private leaderboard introduced an unseen third domain, and the hard routing boundary proved brittle: misclassified questions were irrecoverably directed to the wrong partition, causing a drop from first to fifth place. This confirmed that hard domain routing is a single point of failure whose cost scales with the misclassification rate, and that domain-agnostic retrieval provides the robustness required when hidden-domain generalization is part of the evaluation.

\subsection{Hybrid Retrieval}
We tested sparse retrieval for two purposes: providing broader document context, and exact keyword matching. For the first, sparse signals were consistently outperformed by chunk prepending. For the second, combining dense and sparse signals via rank fusion yielded no gains over dense retrieval alone --- likely because the competition data relies heavily on synonyms and rephrasing, which lexical matching cannot bridge.

\subsection{Agentic Inference}
Finally, we prototyped an agentic solution in which the model interleaves retrieval, reasoning, and query reformulation over multiple steps. The system used a tool-use loop (search $\to$ retrieve pages $\to$ read evidence $\to$ answer) driven by Qwen3.5-4B \cite{qwen35blog}, with hybrid dense+sparse retrieval and an answer repair step. On the competition development set, retrieval navigation was strong (document recall 1.00, page recall 0.96), but final answer accuracy reached only 0.83 --- the small reader model consistently mishandled negation and other fine-grained linguistic cues in Ukrainian answer choices.

Beyond this accuracy gap, embedding a multi-round tool-use agent in a Kaggle code-only competition is not straightforward, so we prioritized a single robust retrieval--reranking--generation pipeline instead.

\section{Conclusions \& Future Work}
\subsection{Summary of Findings}
Our findings can be summarized in four main points. First, we introduced new Ukrainian training resources for retrieval-based multiple-choice document question answering and used them to support both retriever and reranker adaptation. Second, we showed that a strong retrieval-augmented pipeline built largely from modern pretrained components is now sufficient to achieve highly competitive performance for Ukrainian multi-domain document understanding. Third, our embedding experiments revealed a clear efficiency trade-off: larger pretrained embedders already perform very strongly off the shelf, while fine-tuning smaller models can recover much of this gap under tighter compute constraints. Finally, we found that contextualization in chunk construction is one of the most important factors in the entire pipeline, with structure-aware chunking contributing more than many more complex downstream modifications.

Taken together, these results show that strong Ukrainian document QA no longer depends primarily on complex task-specific modeling. Instead, the strongest gains come from combining high-quality pretrained models with careful evidence preparation, efficient retrieval design, and chunk representations that preserve document structure.

\subsection{Future Directions}
Two directions appear especially promising for future work. First, it would be valuable to explore and benchmark diffusion-based and other bidirectional embedding models in a setting without the compute constraints of the shared task, while also investigating practical inference-speed optimizations for such models. Second, late chunking is a highly relevant extension of our current retrieval setup: our results already show that contextualized chunks matter substantially, and late chunking is explicitly designed to give each chunk access to full-document context before pooling, which helps preserve cross-chunk dependencies in retrieval \cite{günther2025latechunkingcontextualchunk}.

\section*{Limitations}
Although the final pipeline is feasible under the Kaggle code-only constraints, it still relies on comparatively large models. This setting rewards fitting the strongest possible components into a fixed offline runtime budget, whereas real production deployments operate under a broader and more dynamic set of practical constraints.

Part of our auxiliary retrieval pretraining corpus was produced by machine translation from English sources. Although we used a high-capacity translation model, the translated examples were not manually reviewed by human annotators.

\section*{Acknowledgements}
We thank the organizers of the UNLP 2026 shared task for preparing the benchmark and the code-only competition setup.

\bibliography{custom}

\begin{thebibliography}{29}
\providecommand{\natexlab}[1]{#1}

\bibitem[{Aarsen(2024)}]{aarsen_natural_questions_hard_negatives}
Tom Aarsen. 2024.
\newblock natural-questions-hard-negatives.
\newblock \url{https://huggingface.co/datasets/tomaarsen/natural-questions-hard-negatives}.
\newblock Hugging Face dataset, accessed 2026-04-08.

\bibitem[{Bazdyrev et~al.(2025)Bazdyrev, Bashtovyi, Havlytskyi, Kharytonov, and Khodakovskyi}]{bazdyrev-etal-2025-transforming}
Anton Bazdyrev, Ivan Bashtovyi, Ivan Havlytskyi, Oleksandr Kharytonov, and Artur Khodakovskyi. 2025.
\newblock \href {https://doi.org/10.18653/v1/2025.unlp-1.13} {Transforming causal {LLM} into {MLM} encoder for detecting social media manipulation in telegram}.
\newblock In \emph{Proceedings of the Fourth Ukrainian Natural Language Processing Workshop (UNLP 2025)}, pages 112--119, Vienna, Austria (online). Association for Computational Linguistics.

\bibitem[{Chen et~al.(2024)Chen, Wang, Chen, Yu, Ma, Zhao, Zhang, and Yu}]{chen-etal-2024-dense}
Tong Chen, Hongwei Wang, Sihao Chen, Wenhao Yu, Kaixin Ma, Xinran Zhao, Hongming Zhang, and Dong Yu. 2024.
\newblock \href {https://doi.org/10.18653/v1/2024.emnlp-main.845} {Dense {X} retrieval: What retrieval granularity should we use?}
\newblock In \emph{Proceedings of the 2024 Conference on Empirical Methods in Natural Language Processing}, pages 15159--15177, Miami, Florida, USA. Association for Computational Linguistics.

\bibitem[{Eslami et~al.(2026)Eslami, Gaiduk, Krimmel, Milliken, Wang, and Bykov}]{eslami2026diffusionpretraineddensecontextualembeddings}
Sedigheh Eslami, Maksim Gaiduk, Markus Krimmel, Louis Milliken, Bo~Wang, and Denis Bykov. 2026.
\newblock \href {https://arxiv.org/abs/2602.11151} {Diffusion-pretrained dense and contextual embeddings}.
\newblock \emph{Preprint}, arXiv:2602.11151.

\bibitem[{Finkelstein et~al.(2026)Finkelstein, Caswell, Domhan, Peter, Juraska, Riley, Deutsch, Kovacs, Dilanni, Cherry, Briakou, Nielsen, Luo, Black, Mullins, Agrawal, Xu, Kats, Jaskiewicz, Freitag, and Vilar}]{translategemma2026}
Mara Finkelstein, Isaac Caswell, Tobias Domhan, Jan-Thorsten Peter, Juraj Juraska, Parker Riley, Daniel Deutsch, Geza Kovacs, Cole Dilanni, Colin Cherry, Eleftheria Briakou, Elizabeth Nielsen, Jiaming Luo, Kat Black, Ryan Mullins, Sweta Agrawal, Wenda Xu, Erin Kats, Stephane Jaskiewicz, and 2 others. 2026.
\newblock \href {https://arxiv.org/abs/2601.09012} {Translate{G}emma technical report}.
\newblock \emph{Preprint}, arXiv:2601.09012.

\bibitem[{Günther et~al.(2025)Günther, Mohr, Williams, Wang, and Xiao}]{günther2025latechunkingcontextualchunk}
Michael Günther, Isabelle Mohr, Daniel~James Williams, Bo~Wang, and Han Xiao. 2025.
\newblock \href {https://arxiv.org/abs/2409.04701} {Late chunking: Contextual chunk embeddings using long-context embedding models}.
\newblock \emph{Preprint}, arXiv:2409.04701.

\bibitem[{Haltiuk and Smywi{\'n}ski-Pohl(2025)}]{haltiuk-smywinski-pohl-2025-path}
Mykola Haltiuk and Aleksander Smywi{\'n}ski-Pohl. 2025.
\newblock \href {https://doi.org/10.18653/v1/2025.unlp-1.14} {On the path to make {U}krainian a high-resource language}.
\newblock In \emph{Proceedings of the Fourth Ukrainian Natural Language Processing Workshop (UNLP 2025)}, pages 120--130, Vienna, Austria (online). Association for Computational Linguistics.

\bibitem[{Han and Han(2025)}]{han2025gemma3}
Daniel Han and Michael Han. 2025.
\newblock \href {https://unsloth.ai/blog/gemma3} {Fine-tune \& run gemma 3}.

\bibitem[{Ivanyuk-Skulskiy et~al.(2021)Ivanyuk-Skulskiy, Zaliznyi, Reshetar, Protsyk, Romanchuk, and Shpihanovych}]{ua_datasets_2021}
Bogdan Ivanyuk-Skulskiy, Anton Zaliznyi, Oleksand Reshetar, Oleksiy Protsyk, Bohdan Romanchuk, and Vladyslav Shpihanovych. 2021.
\newblock \href {https://github.com/fido-ai/ua-datasets} {ua\_datasets: a collection of ukrainian language datasets}.

\bibitem[{Khashabi et~al.(2021)Khashabi, Ng, Khot, Sabharwal, Hajishirzi, and Callison-Burch}]{gooaq2021}
Daniel Khashabi, Amos Ng, Tushar Khot, Ashish Sabharwal, Hannaneh Hajishirzi, and Chris Callison-Burch. 2021.
\newblock Gooaq: Open question answering with diverse answer types.
\newblock \emph{arXiv preprint}.

\bibitem[{Limkonchotiwat et~al.(2024)Limkonchotiwat, Ponwitayarat, Lowphansirikul, Manakul, Udomcharoenchaikit, Chuangsuwanich, and Nutanong}]{limkonchotiwat-etal-2024-mccrolin}
Peerat Limkonchotiwat, Wuttikorn Ponwitayarat, Lalita Lowphansirikul, Potsawee Manakul, Can Udomcharoenchaikit, Ekapol Chuangsuwanich, and Sarana Nutanong. 2024.
\newblock \href {https://doi.org/10.18653/v1/2024.findings-emnlp.157} {{M}c{C}rolin: Multi-consistency cross-lingual training for retrieval question answering}.
\newblock In \emph{Findings of the Association for Computational Linguistics: EMNLP 2024}, pages 2780--2793, Miami, Florida, USA. Association for Computational Linguistics.

\bibitem[{Lin(2024)}]{lin2024chatdoc}
Demiao Lin. 2024.
\newblock \href {https://arxiv.org/abs/2401.12599} {Revolutionizing retrieval-augmented generation with enhanced pdf structure recognition}.
\newblock \emph{Preprint}, arXiv:2401.12599.

\bibitem[{Lin et~al.(2024)Lin, Tang, Tang, Yang, Chen, Wang, Xiao, Dang, Gan, and Han}]{lin2024awqactivationawareweightquantization}
Ji~Lin, Jiaming Tang, Haotian Tang, Shang Yang, Wei-Ming Chen, Wei-Chen Wang, Guangxuan Xiao, Xingyu Dang, Chuang Gan, and Song Han. 2024.
\newblock \href {https://arxiv.org/abs/2306.00978} {Awq: Activation-aware weight quantization for llm compression and acceleration}.
\newblock \emph{Preprint}, arXiv:2306.00978.

\bibitem[{Lu et~al.(2025)Lu, Chen, Qiao, and Sun}]{hichunk2025}
Wensheng Lu, Keyu Chen, Ruizhi Qiao, and Xing Sun. 2025.
\newblock \href {https://arxiv.org/abs/2509.11552} {Hichunk: Evaluating and enhancing retrieval-augmented generation with hierarchical chunking}.
\newblock \emph{Preprint}, arXiv:2509.11552.

\bibitem[{Merola and Singh(2025)}]{singh2025contextual}
Carlo Merola and Jaspinder Singh. 2025.
\newblock \href {https://arxiv.org/abs/2504.19754} {Reconstructing context: Evaluating advanced chunking strategies for retrieval-augmented generation}.
\newblock \emph{Preprint}, arXiv:2504.19754.

\bibitem[{Nguyen et~al.(2016)Nguyen, Rosenberg, Song, Gao, Tiwary, Majumder, and Deng}]{DBLP:journals/corr/NguyenRSGTMD16}
Tri Nguyen, Mir Rosenberg, Xia Song, Jianfeng Gao, Saurabh Tiwary, Rangan Majumder, and Li~Deng. 2016.
\newblock \href {https://arxiv.org/abs/1611.09268} {{MS} {MARCO:} {A} human generated machine reading comprehension dataset}.
\newblock \emph{CoRR}, abs/1611.09268.

\bibitem[{Paniv et~al.(2025)Paniv, Didenko, Haltiuk, Humennyy, Kravchenko, Kyslyi, Makovska, Orlovskyi, Ruban, Rudko, Senyk, Drushchak, Chaplynskyi, and Romanyshyn}]{Paniv_Lapa_LLM_v0_1_2_2025}
Yurii Paniv, Bohdan Didenko, Mykola Haltiuk, Vladyslav Humennyy, Andrian Kravchenko, Roman Kyslyi, Viktoriia Makovska, Artem Orlovskyi, Bohdan Ruban, Maksym-Yurii Rudko, Anastasiia Senyk, Nazarii Drushchak, Dmytro Chaplynskyi, and Mariana Romanyshyn. 2025.
\newblock \href {https://github.com/lapa-llm/lapa-llm/} {{Lapa LLM v0.1.2 — the most efficient Ukrainian open-source language model}}.

\bibitem[{Rajpurkar et~al.(2018)Rajpurkar, Jia, and Liang}]{rajpurkar-etal-2018-know}
Pranav Rajpurkar, Robin Jia, and Percy Liang. 2018.
\newblock \href {https://doi.org/10.18653/v1/P18-2124} {Know what you don{'}t know: Unanswerable questions for {SQ}u{AD}}.
\newblock In \emph{Proceedings of the 56th Annual Meeting of the Association for Computational Linguistics (Volume 2: Short Papers)}, pages 784--789, Melbourne, Australia. Association for Computational Linguistics.

\bibitem[{Rajpurkar et~al.(2016)Rajpurkar, Zhang, Lopyrev, and Liang}]{rajpurkar-etal-2016-squad}
Pranav Rajpurkar, Jian Zhang, Konstantin Lopyrev, and Percy Liang. 2016.
\newblock \href {https://doi.org/10.18653/v1/D16-1264} {{SQ}u{AD}: 100,000+ questions for machine comprehension of text}.
\newblock In \emph{Proceedings of the 2016 Conference on Empirical Methods in Natural Language Processing}, pages 2383--2392, Austin, Texas. Association for Computational Linguistics.

\bibitem[{Sydorskyi et~al.(2026)Sydorskyi, Romanyshyn, Kyslyi, and Nahorna}]{sydorskyi-etal-2026-unlp}
Volodymyr Sydorskyi, Nataliia Romanyshyn, Roman Kyslyi, and Olena Nahorna. 2026.
\newblock The {UNLP} 2026 shared task on multi-domain document understanding.
\newblock In \emph{Proceedings of the Fifth Ukrainian Natural Language Processing Conference (UNLP 2026)}, Lviv, Ukraine. Association for Computational Linguistics.
\newblock To appear.

\bibitem[{Team(2024)}]{Docling}
Deep~Search Team. 2024.
\newblock \href {https://doi.org/10.48550/arXiv.2408.09869} {Docling technical report}.
\newblock Technical report.

\bibitem[{Team et~al.(2025)Team, Kamath, Ferret, Pathak, Vieillard, Merhej, Perrin, Matejovicova, Ramé, Rivière, Rouillard, Mesnard, Cideron, bastien Grill, Ramos, Yvinec, Casbon, Pot, Penchev, Liu, Visin, Kenealy, Beyer, Zhai, Tsitsulin, Busa-Fekete, Feng, Sachdeva, Coleman, Gao, Mustafa, Barr, Parisotto, Tian, Eyal, Cherry, Peter, Sinopalnikov, Bhupatiraju, Agarwal, Kazemi, Malkin, Kumar, Vilar, Brusilovsky, Luo, Steiner, Friesen, Sharma, Sharma, Gilady, Goedeckemeyer, Saade, Feng, Kolesnikov, Bendebury, Abdagic, Vadi, György, Pinto, Das, Bapna, Miech, Yang, Paterson, Shenoy, Chakrabarti, Piot, Wu, Shahriari, Petrini, Chen, Lan, Choquette-Choo, Carey, Brick, Deutsch, Eisenbud, Cattle, Cheng, Paparas, Sreepathihalli, Reid, Tran, Zelle, Noland, Huizenga, Kharitonov, Liu, Amirkhanyan, Cameron, Hashemi, Klimczak-Plucińska, Singh, Mehta, Lehri, Hazimeh, Ballantyne, Szpektor, Nardini, Pouget-Abadie, Chan, Stanton, Wieting, Lai, Orbay, Fernandez, Newlan, yeong Ji, Singh, Black, Yu, Hui, Vodrahalli, Greff, Qiu,
  Valentine, Coelho, Ritter, Hoffman, Watson, Chaturvedi, Moynihan, Ma, Babar, Noy, Byrd, Roy, Momchev, Chauhan, Sachdeva, Bunyan, Botarda, Caron, Rubenstein, Culliton, Schmid, Sessa, Xu, Stanczyk, Tafti, Shivanna, Wu, Pan, Rokni, Willoughby, Vallu, Mullins, Jerome, Smoot, Girgin, Iqbal, Reddy, Sheth, Põder, Bhatnagar, Panyam, Eiger, Zhang, Liu, Yacovone, Liechty, Kalra, Evci, Misra, Roseberry, Feinberg, Kolesnikov, Han, Kwon, Chen, Chow, Zhu, Wei, Egyed, Cotruta, Giang, Kirk, Rao, Black, Babar, Lo, Moreira, Martins, Sanseviero, Gonzalez, Gleicher, Warkentin, Mirrokni, Senter, Collins, Barral, Ghahramani, Hadsell, Matias, Sculley, Petrov, Fiedel, Shazeer, Vinyals, Dean, Hassabis, Kavukcuoglu, Farabet, Buchatskaya, Alayrac, Anil, Dmitry, Lepikhin, Borgeaud, Bachem, Joulin, Andreev, Hardin, Dadashi, and Hussenot}]{gemmateam2025gemma3technicalreport}
Gemma Team, Aishwarya Kamath, Johan Ferret, Shreya Pathak, Nino Vieillard, Ramona Merhej, Sarah Perrin, Tatiana Matejovicova, Alexandre Ramé, Morgane Rivière, Louis Rouillard, Thomas Mesnard, Geoffrey Cideron, Jean bastien Grill, Sabela Ramos, Edouard Yvinec, Michelle Casbon, Etienne Pot, Ivo Penchev, and 197 others. 2025.
\newblock \href {https://arxiv.org/abs/2503.19786} {Gemma 3 technical report}.
\newblock \emph{Preprint}, arXiv:2503.19786.

\bibitem[{Team(2026)}]{qwen35blog}
Qwen Team. 2026.
\newblock \href {https://qwen.ai/blog?id=qwen3.5} {Qwen3.5: Accelerating productivity with native multimodal agents}.

\bibitem[{Wang et~al.(2024)Wang, Huang, Jackson, and Gao}]{wang-etal-2024-retrieve}
Dingmin Wang, Qiuyuan Huang, Matthew Jackson, and Jianfeng Gao. 2024.
\newblock \href {https://doi.org/10.1162/tacl_a_00646} {Retrieve what you need: A mutual learning framework for open-domain question answering}.
\newblock \emph{Transactions of the Association for Computational Linguistics}, 12:247--263.

\bibitem[{Wang et~al.(2025)Wang, Gao, Xiao, Huang, Si, Luo, Bai, Li, Duan, Lv, Lu, Chen, Qi, and Sun}]{wang-etal-2025-document}
Zhitong Wang, Cheng Gao, Chaojun Xiao, Yufei Huang, Shuzheng Si, Kangyang Luo, Yuzhuo Bai, Wenhao Li, Tangjian Duan, Chuancheng Lv, Guoshan Lu, Gang Chen, Fanchao Qi, and Maosong Sun. 2025.
\newblock \href {https://doi.org/10.18653/v1/2025.findings-acl.422} {Document segmentation matters for retrieval-augmented generation}.
\newblock In \emph{Findings of the Association for Computational Linguistics: ACL 2025}, pages 8063--8075, Vienna, Austria. Association for Computational Linguistics.

\bibitem[{Yang et~al.(2025)Yang, Li, Yang, Zhang, Hui, Zheng, Yu, Gao, Huang, Lv, Zheng, Liu, Zhou, Huang, Hu, Ge, Wei, Lin, Tang, Yang, Tu, Zhang, Yang, Yang, Zhou, Zhou, Lin, Dang, Bao, Yang, Yu, Deng, Li, Xue, Li, Zhang, Wang, Zhu, Men, Gao, Liu, Luo, Li, Tang, Yin, Ren, Wang, Zhang, Ren, Fan, Su, Zhang, Zhang, Wan, Liu, Wang, Cui, Zhang, Zhou, and Qiu}]{yang2025qwen3technicalreport}
An~Yang, Anfeng Li, Baosong Yang, Beichen Zhang, Binyuan Hui, Bo~Zheng, Bowen Yu, Chang Gao, Chengen Huang, Chenxu Lv, Chujie Zheng, Dayiheng Liu, Fan Zhou, Fei Huang, Feng Hu, Hao Ge, Haoran Wei, Huan Lin, Jialong Tang, and 41 others. 2025.
\newblock \href {https://arxiv.org/abs/2505.09388} {Qwen3 technical report}.
\newblock \emph{Preprint}, arXiv:2505.09388.

\bibitem[{Yang et~al.(2018)Yang, Qi, Zhang, Bengio, Cohen, Salakhutdinov, and Manning}]{yang2018hotpotqa}
Zhilin Yang, Peng Qi, Saizheng Zhang, Yoshua Bengio, William Cohen, Ruslan Salakhutdinov, and Christopher~D. Manning. 2018.
\newblock \href {https://doi.org/10.18653/v1/D18-1259} {{H}otpot{QA}: A dataset for diverse, explainable multi-hop question answering}.
\newblock In \emph{Proceedings of the 2018 Conference on Empirical Methods in Natural Language Processing}, pages 2369--2380, Brussels, Belgium. Association for Computational Linguistics.

\bibitem[{Yukhymenko et~al.(2025)Yukhymenko, Alexandrov, and Vechev}]{MamayLMv1}
Hanna Yukhymenko, Anton Alexandrov, and Martin Vechev. 2025.
\newblock Mamaylm v1.0: An efficient state-of-the-art multimodal ukrainian llm.

\bibitem[{Zhang et~al.(2025)Zhang, Li, Long, Zhang, Lin, Yang, Xie, Yang, Liu, Lin, Huang, and Zhou}]{zhang2025qwen3embeddingadvancingtext}
Yanzhao Zhang, Mingxin Li, Dingkun Long, Xin Zhang, Huan Lin, Baosong Yang, Pengjun Xie, An~Yang, Dayiheng Liu, Junyang Lin, Fei Huang, and Jingren Zhou. 2025.
\newblock \href {https://arxiv.org/abs/2506.05176} {Qwen3 embedding: Advancing text embedding and reranking through foundation models}.
\newblock \emph{Preprint}, arXiv:2506.05176.

\end{thebibliography}

\appendix

\section{Retrieval Prompt}
\label{fst:appendix}
\begin{Prompt}
Instruct: Given a multiple-choice question in Ukrainian, retrieve relevant passages from Ukrainian PDF documents that help identify the correct supporting document and page.
Query: {question}
Options:
{choices}
\end{Prompt}

\section{Reranking Prompts}
\label{sec:appendix}
\subsection*{System Prompt}
\label{app:rerank-system-prompt}

\begin{Prompt}
Judge whether the Document meets the requirements based on the Query and the Instruct provided. Note that the answer can only be "yes" or "no".
\end{Prompt}

\subsection*{User Prompt Template}
\label{app:rerank-user-prompt}
\promptlabel{Original}
\begin{otherlanguage*}{ukrainian}
\begin{Prompt}
<Instruct>: Given a web search query, retrieve relevant passages that answer the query.
<Query>: {question}
Варіанти відповідей:
{choices}

<Document>: {document}
\end{Prompt}
\end{otherlanguage*}

\promptlabel{English translation}
\begin{Prompt}
<Instruct>: Given a web search query, retrieve relevant passages that answer the query.
<Query>: {question}
Answer options:
{choices}

<Document>: {document}
\end{Prompt}

\section{Answer Generation Prompts}
\label{trd:appendix}
\subsection*{System Prompt}
\label{app:system-prompt}

\promptlabel{Original}
\begin{otherlanguage*}{ukrainian}
\begin{Prompt}
Ти розв'язуєш завдання множинного вибору за наданими уривками з документів. Уважно прочитай запитання, варіанти відповідей і всі уривки. Менший ранг пошуку означає сильніший сигнал релевантності. Якщо уривки суперечать один одному, віддавай перевагу більш прямому та конкретному формулюванню. Якщо інформації недостатньо, все одно обери найімовірніший варіант. Поверни лише одну велику латинську літеру: A, B, C, D, E або F.
\end{Prompt}
\end{otherlanguage*}

\promptlabel{English translation}
\begin{Prompt}
You are solving a multiple-choice task using the provided document excerpts. Carefully read the question, answer options, and all excerpts. A lower retrieval rank indicates a stronger relevance signal. If the excerpts contradict each other, prefer the more direct and specific formulation. If the information is insufficient, still choose the most likely option. Return only one uppercase Latin letter: A, B, C, D, E, or F.
\end{Prompt}

\subsubsection*{User Prompt Template}
\label{app:user-prompt}
\promptlabel{Original}
\begin{otherlanguage*}{ukrainian}
\begin{Prompt}
Запитання: {question}

Варіанти відповідей:
{choices}

Надані уривки (менший ранг пошуку = сильніший сигнал):
{context_blocks}

Запитання: {question}

Варіанти відповідей:
{choices}

Відповідь (лише одна літера A-F):
\end{Prompt}
\end{otherlanguage*}

\promptlabel{English Translation}
\begin{Prompt}
Question: {question}

Answer options:
{choices}

Provided excerpts (lower retrieval rank = stronger relevance signal):
{context_blocks}

Question: {question}

Answer options:
{choices}

Answer (only one letter A-F):
\end{Prompt}

\end{document}